%% file: main.tex
\pdfoutput=1

\documentclass[11pt]{article}

\usepackage{acl}

\usepackage{times}
\usepackage{latexsym}
\usepackage[T1]{fontenc}
\usepackage[utf8]{inputenc}
\usepackage{microtype}
\usepackage{inconsolata}
\usepackage{graphicx}
\usepackage{caption}
\usepackage{amsmath}
\usepackage{amssymb}
\usepackage{comment}
\usepackage{booktabs}
\usepackage{hyperref}
\usepackage[overload]{empheq}
\usepackage{makecell}
\usepackage{xcolor,colortbl}
\usepackage[ruled,vlined]{algorithm2e}
\usepackage{array,multirow}
\usepackage{comment}
\usepackage{pifont}
\usepackage{siunitx}
\usepackage{url}
\usepackage{enumitem}
\usepackage{arydshln}
\usepackage{xcolor}
\usepackage{soul}

\newcommand{\hlc}[2][yellow]{{%
    \colorlet{foo}{#1}%
    \sethlcolor{foo}\hl{#2}}%
}

\newcommand{\cmark}{\ding{51}}%
\newcommand{\xmark}{\ding{55}}%

\newcommand{\nth}[1]{#1^{\text{th}}}
\newcommand{\CoHfunc}{
\textrm{H}
}
\newcommand{\CoH}[2]{
\CoHfunc(#1,\:#2)
}
\newcommand{\NCoHfunc}{
\tilde{\textrm{H}}
}
\newcommand{\NCoH}[2]{
\NCoHfunc(#1,\:#2)
}
\newcommand{\ANCoHfunc}{
\textrm{A}\tilde{\textrm{H}}
}

\title{A Novel Metric for Measuring the Robustness of Large Language Models \\in Non-adversarial Scenarios}

\author{
	Samuel Ackerman\hspace{1cm}
	Ella Rabinovich\hspace{1cm}
	\textbf{Eitan Farchi}\hspace{1cm} 
	\textbf{Ateret Anaby-Tavor} 
	\vspace{0.15cm} \\
	IBM Research \\
	\texttt{\{samuel.ackerman, ella.rabinovich1\}@ibm.com} \\
	\texttt{\{farchi, atereta\}@il.ibm.com}
}

\begin{document}
\maketitle
\begin{abstract}
We evaluate the robustness of several large language models on multiple datasets. Robustness here refers to the relative insensitivity of the model's answers to meaning-preserving variants of their input. Benchmark datasets are constructed by introducing naturally-occurring, non-malicious perturbations, or by generating semantically equivalent paraphrases of input questions or statements. We further propose a novel metric for assessing a model robustness, and demonstrate its benefits in the non-adversarial scenario by empirical evaluation of several models on the created datasets.
\footnote{All our datasets are available on \href{https://huggingface.co/collections/ibm/paraphrase-and-perturbation-question-answering-robustness-66c314dc70eace5a99f15a63}{HuggingFace}. 
Additionally, a dataset generation results and jupyter notebooks to produce images for this paper are available at \url{https://github.com/IBM/EMNLP_2024_LLM_robustness/tree/master}.}

\end{abstract}

\input{chapters/introduction-related-work}
\input{chapters/datasets}
\input{chapters/benchmarking-definitions}
\input{chapters/benchmarking-results}
\input{chapters/conclusion}
\input{chapters/limitations}


\bibliography{main}

\input{chapters/appendix}

\end{document}

%% file: chapters/introduction-related-work.tex
\section{Introduction}
\label{sec:introduction}

With the increase in the prominence and use of large language models (LLMs), there has been tremendous activity in evaluating various aspects of these models' behavior and its alignment with desirable qualities, such as accuracy, safety and privacy. The property of model \textit{robustness}---the ability of a model to produce semantically equivalent output given meaning-preserving input---has been addressed from various perspectives: sensitivity to the wording of instruction template \citep{mizrahi2023stateofwhat, sclar2023quantifying, zhao2024improving}, example choice and ordering for in-context learning (ICL) tasks \citep{voronov2024mind}, perturbing the order of premises in logical reasoning tasks \citep{chen2024premise}, as well as model resilience to adversarial prompts (e.g., \citet{liang2023holistic, zhou2022prompt, shayegani2023survey}).

Model robustness (or insensitivity) to naturally-occurring, non-malicious variations in their input, such as the question in a question-answering (QA) task, or the statement in a classification task, has received relatively little attention (although see \citet{liang2023holistic} for sensitivity to typos, and \citet{raj2022measuring} for evaluation of model consistency on input paraphrases). Such slight variations include perturbations that can normally occur in human-generated input, e.g., changes in casing, redundant white-spacing or newlines, the lack of punctuation, "butter-finger" typos, 
character swap, and meaning preserving paraphrases. While very common in everyday language, these changes may have a significant effect on a model's ability to produce anticipated answers. The reading comprehension example in Table~\ref{tbl:robustness-example} illustrates how slight changes in the phrasing of a question in one of our datasets cause the model to generate different responses. 

\begin{table*}[h!]
\centering
\resizebox{\textwidth}{!}{
\begin{tabular}{p{16.75cm}}
\textbf{Context:} A function f is said to be continuously differentiable if the derivative f'(x) exists and is itself a continuous function. Though the derivative of a differentiable function never has a jump discontinuity, 
it is possible for the derivative to have an essential discontinuity. For example, the function [...] \\ \hline \hline
\end{tabular}
}
\resizebox{\textwidth}{!}{
\begin{tabular}{p{14cm}|c|c}
question & model answer & correct \\ \hline
(1) is the derivative of a continuous function always continuous? & yes & \xmark \\ \hdashline
(2) \hlc[yellow!50]{Is} the derivative of a continuous function always continuous & no & \cmark \\
(3) Is the \hspace{0.5cm} derivative of a continuous function \hspace{0.5cm} always continuous? & yes & \xmark \\
(4) IS THE DERIVATIVE OF A CONTINUOUS FUNCTION ALWAYS CONTINUOUS? & no & \cmark \\
(5) \hlc[yellow!50]{does} the derivative of a continuous function \hlc[yellow!50]{always exhibit continuous behavior?} & yes & \xmark \\ 
(6) is the derivative of a continuous function \hlc[yellow!50]{guaranteed to be continuous?} & yes & \xmark \\
\end{tabular}
}
\vspace{-0.05in}
\caption{Llama2-chat (13B) model's answers to the original question (1) and its perturbed variants (2-6), where simple superficial perturbations were applied to obtain variants (2-4), and a paraphrasing model produced variants (5-6). The LLM's answer to the original question is incorrect, two of the variants --- (2) and (4) --- obtained the correct answer ("no"). Variants' distinctions from the original phrasing are highlighted, where not easily visible.}
\label{tbl:robustness-example}
\end{table*}

Benchmarking the robustness of a model to variations in its input typically involves measuring the degree of performance decrease in the perturbed instance set, compared to the original example. For assessing the resilience of LLMs to \textit{adversarial attacks}, the main metric that has been put forward in prior studies is \textit{performance drop rate} (PDR), which is the fractional decrease in the average perturbed instances' score, relative to the original example \citep{zhu2023promptbench}. As discussed in Section~\ref{sssec:pdr}, PDR has two main drawbacks: First, since it measures fractional \emph{decrease}, it is inherently an \emph{asymmetric} function of its inputs; a fixed increase in performance after perturbation receives a larger magnitude PDR than the reversed decrease in performance. Second, since fractional change from 0 is undefined, the PDR is \textit{undefined} in the specific case when the original score was zero but the average over perturbed instance set scores higher, as in the example in Table~\ref{tbl:robustness-example}; thus, instances with undefined PDR are ignored when evaluating average PDR on a dataset (e.g., \citet{liang2023holistic, zhu2023promptbench}), which can bias aggregates.

While performance improvement is not typical to adversarial tests, it can easily happen in the scenario with naturally-occurring, non-malicious input variations which we consider. Moreover, scoring a model output with 0 is not uncommon in tasks with binary-valued evaluation (correct or wrong), as in our study.
%
%
Aiming to overcome these drawbacks, we adapt the Cohen's $h$ statistical effect size metric \cite{C1988} for the difference in proportions, discussed in Section~\ref{sssec:cohens_h_effect_size}.  
Indicating the practical significance of a difference in two groups (e.g., outcome of two experimental settings), the use of effect sizes is widely practiced in research and commercial applications (see \citet{fritz2012effect}). We show that in the setting of robustness evaluation, Cohen's $h$ constitutes an elegant, symmetric and easily-interpretable metric, which correlates with PDR while overcoming its drawbacks.


Our contribution is, therefore, twofold: First, we expand multiple datasets, concerning classification, QA and reading comprehension, with naturally-occurring input variants, and report a comprehensive assessment of the  robustness of LLMs on these tasks. Second, we propose, and empirically evaluate, a novel metric for measuring model sensitivity to non-adversarial perturbations in its input.
Much effort has been invested recently in leaderboards for multi-faceted evaluation of foundation models (e.g., \citet{liang2023holistic}\footnote{\url{https://crfm.stanford.edu/helm/}}). A broader impact of this study lies in the adoption of the proposed metric for benchmarking the robustness of LLMs in non-adversarial scenarios.

%% file: chapters/datasets.tex
\section{Datasets}
\label{sec:dataset}

\subsection{Dataset Description}
\label{ssec:dataset_description}

We make use of multiple diverse datasets in our experiments. The original datasets are expanded by introducing various types of perturbations into raw instances: superficial (non-semantic), paraphrasing, and adding distraction passages where applicable. We experiment with three datasets: (1) PopQA \citep{mallen2023not}: open-domain questions of factual nature about public figures and entities (books, countries, etc.); the dataset has been recently expanded with manually-generated paraphrases by \citet{rabinovich2023predicting}; (2) social identity group abuse (SIGA): short statements for classification, that possibly carry over an abusive flavor towards an identity group by race, religion or gender \citep{wiegand2022identifying}; 
(3) BoolQ: a dataset of reading-comprehension questions, with boolean answers \citep{clark2019boolq}. We use \textit{string containment} as the evaluation metric on PopQA (as in \citet{mallen2023not}), and \textit{accuracy} for BoolQ and SIGA.

\subsection{Expanding Datasets with Perturbations}
\label{ssec:input_perturbations}
We imitate naturally-occurring variations in human-generated input, by applying the following perturbation types on each input in the three datasets:

\vspace{-0.05in}
\paragraph{Superficial (S)} Simple non-semantic perturbations, such as upper-, lower- or proper-casing of certain words, removing punctuation, "butter-finger" typos (misspelling by replacing a randomly-selected letter with one of the adjacent ones on a keyboard), 
character swap, or redundant white-spacing. A sentence variant can include one or multiple interventions from this set, as illustrated in examples (2)-(4) in Table~\ref{tbl:robustness-example}.

\vspace{-0.05in}
\paragraph{Paraphrase (P)} We automatically generate (at most) five semantics-preserving paraphrases using the NL-Augmenter \cite{dhole2023nl}
paraphrase generator.
For PopQA, we use the manually-crafted templated paraphrases by \citet{rabinovich2023predicting}.

\vspace{-0.05in}
\paragraph{Distraction (D)} BoolQ --- the reading comprehension dataset --- was additionally expanded with "distractions": a randomly selected passage from the corpus was appended before or after the passage in the input example, to assess models' resilience to (possibly related but not strictly relevant) information in the content-grounded QA task.

Table~\ref{tbl:datasets_stats} summarizes the datasets statistics before and after expansion, and perturbation types used. Let $\mathcal D$ denote an unperturbed test dataset, consisting of $n$ unique instances $x_1,\dots,x_n$ (e.g., questions to be answered). The expanded dataset $\mathcal D'$ consists of each \textit{original} instance $x_i$ (now denoted by $x_i^o$), as well as the set of its $m(i){\geq}1$ perturbations, denoted by $(x_i^1,\dots,x_i^{m(i)})$.  Our expanded datasets are available on \href{https://huggingface.co/collections/ibm/paraphrase-and-perturbation-question-answering-robustness-66c314dc70eace5a99f15a63}{HuggingFace}. 


\begin{table}[h!]
\centering
\resizebox{1.0\columnwidth}{!}{
\begin{tabular}{l|r|rrr|r} 
dataset & original & S & P &	D & final \\  \hline
PopQA   & 14.2K & 85.6K & 104.3K    & --    & 204.2K \\
BoolQ   & 3.2K  & 9.8K  & 9.7K      & 6.5K  & 29.3K \\
SIGA    & 2.1K  & 12.6K & 6.1K      & --    & 20.8K \\
\end{tabular}
}
\caption{Datasets size before and after expansion, by perturbation type: superficial (S), paraphrase (P) and distraction (D). Only test set portion of the dataset was considered for experiments where applicable.}  
\label{tbl:datasets_stats}
\end{table}

%% file: chapters/benchmarking-definitions.tex
\section{Quantifying Model Robustness}
\label{sec:benchmarking-experiments}


Consider $x_i^o$ and $(x_i^1,\dots,x_i^{m(i)})$, the $\nth{i}$ original input and its $m(i)$ perturbations, in the perturbed dataset $\mathcal D'$ (see Section~\ref{ssec:input_perturbations}).  Given a scoring metric $score\in[0,1]$, let $score_i^o$ and $(score_i^1,\dots,score_i^{m(i)})$ be scores of the model's predicted value (e.g., generated answer) on the original ("o") and perturbed  instances, respectively, versus the ground truth reference.  In our case, $score\in\{0,1\}$ are binary-valued accuracy or string containment match metrics.
We compute the average score on the perturbed ("p") instance set for input $x_i^o$ in $\mathcal D'$ by $score_i^p=\frac{1}{m(i)}\sum_{j=1}^{m(i)}score_i^j$.
We consider a model's performance on a dataset as being \textit{robust} if the performance tends to be insensitive to perturbations; that is, the two scores---$score_i^o$ and $score_i^p$---tend to be close to to each other, across all input instances $i$ in $\mathcal D$.

Note that the notion of model robustness differs from the model's performance overall, which would assess the averages of the scores (either original, perturbed, or both). An LLM can be robust but have poor performance, or have high average performance on the original instances ($score_i^o$) but perform poorly on perturbations ($score_i^p$), making it sensitive to variation in its input.
We now describe the two metrics used to measure model performance robustness: performance drop rate (PDR) \citep{zhu2023promptbench}, traditionally used to assess an LLM's resilience to adversarial attacks, and Cohen's $h$ effect size -- the proposed metric for assessing models' robustness to naturally-occurring, non-malicious perturbations. 

\subsection{Performance Drop Rate (PDR)
\label{sssec:pdr}}

PDR \citep{zhu2023promptbench}, the fractional change in the mean perturbed score of example $i$, relative to the original, is defined\footnote{We added the first case to the definition in \citet{zhu2023promptbench} for scenarios where both the original and the perturbed instances' performance are incorrect.} as PDR($score_i^o$,$\:score_i^p$) =
\[
\begin{cases}
0, & score_i^o=score_i^p=0 \\
\textrm{undefined}, & score_i^o=0, score_i^p \ne 0 \\
1-\frac{score_i^p}{score_i^o}, & \textrm{otherwise}
\end{cases}
\]


Due to its asymmetric nature, PDR is biased towards cases where $score_i^p{>}score_i^o$, contrary to the opposite scenarios, skewing the final average score. As a concrete example, the increase from 0.1 (original) to 0.8 (perturbed set) has a PDR of -7 (=-700\%), while the opposite direction, a decrease from 0.8 (original) to 0.1 (perturbed set), has a PDR of 0.875 (=87.5\%).
Additionally, the metric is udenfined in cases where the model's performance on the original instance is incorrect ($score_i^o{=}0$).
Collectively, these characteristics make PDR a sub-optimal choice for assessing a model robustness to perturbations in non-adversarial scenarios.\footnote{Adaptations of PDR addressing (to some extent) its drawbacks can be devised, but they harm the semantics of PDR, being defined as performance drop ratio.}

\subsection{Cohen's $h$ Effect Size}
\label{sssec:cohens_h_effect_size}

Cohen's $h$ \cite{C1988} statistical effect size is commonly used for measuring the difference in two proportions in empirical research (see Appendix~\ref{sec:appendix_effect_size} for background), and is defined as
\[
\begin{array}{c}
\CoH{score_i^o}{score_i^p}=\psi({score_i^p})-\psi({score_i^o}), \\
\text{where  } \psi({score_i})=2\left(\arcsin{(\sqrt{score_i})}\right)
\end{array}
\]

Cohen's $h$ effect size takes values in the range $[-\pi, \pi]$, where $h{>}0$ indicates performance improvement relative to $score_i^o$. This metric has several important characteristics: (1) Unlike PDR, it is a symmetric function, and is defined for all pairs of $score$ values within the [0, 1] range.
(2) It has rule-of-thumb thresholds\footnote{See Table~\ref{tab:effect_size_thresholds} in Appendix~\ref{sec:appendix_effect_size}.} of what values constitute small, medium, etc. differences in sample proportions, which are based on its statistical properties. For better interpretability, we define a normalized version of Cohen's $h$, defined as $\NCoHfunc=\CoHfunc/\pi$, which takes values within the [-1, 1] range. Consequently, the effect size thresholds are adjusted for this normalized version, each divided by $\pi$.  

\begin{figure}[h!]
\centering
\includegraphics[width=\linewidth]{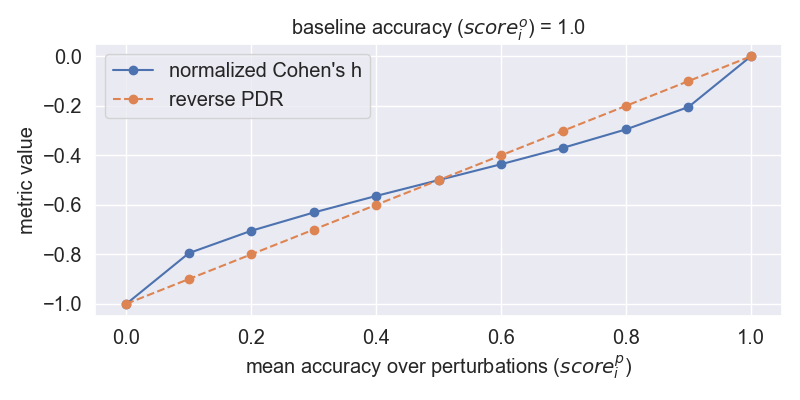}
\caption{Comparison of normalized Cohen's $h$ ($\NCoHfunc$) and reverse PDR (=$-1{\times}\textrm{PDR}$) when the original instance accuracy $score_i^o{=}1.0$ (as in the tasks in our study -- binary evaluation outcome: 0 or 1).}
\label{fig:pdr_cohensh}
\end{figure}

Figure~\ref{fig:pdr_cohensh} shows that PDR and $\NCoHfunc$ correlate very well (Pearson's $r{\approx}0.995$ when $score_i^o{=}1.0$), supporting this novel application of the metric to the scenario of tasks with binary-valued evaluation outcome. We note, however, that the application of the metric is not limited to the scenario with binary evaluation outcome. See Appendix~\ref{sec:appendix_effect_size} for a statistical discussion of the $\NCoHfunc$ metric. 

The absolute value of a directional effect sizes ($\ANCoHfunc=|\NCoHfunc|$) measures the degree of deviation in either direction, and can serve as a proxy for the expected \textit{variance} or absolute deviation between the original and perturbed instance performance.  

Taking the example values of $score_i^o=0.8$ and $score_i^p=0.1$ used above for PDR (Section~\ref{sssec:pdr}), the corresponding value of $\NCoHfunc$ is -0.5, and 0.5 if the direction is reversed (non-normalized $\CoHfunc\approx \pm 1.57$).  This counts as a `very large' difference according to the thresholds in Table~\ref{tab:effect_size_thresholds} (see Appendix~\ref{sec:appendix_effect_size}).

%% file: chapters/benchmarking-results.tex
\section{Benchmarking Model Robustness}

\subsection{Experimental Setup}
We conduct experiments using the following LLMs, proven effective in multiple tasks: instruction-tuned Google's Flan-T5-XXL (11B; \citet{chung2022scaling}) and Flan-UL2 (20B; \citet{flan_ul2}), IBM's Granite 13B series: Chat and Instruct \cite{granite}, Meta AI's Llama2-Chat (13B; \citet{touvron2023llama}) and the recent Llama3-Instruct (70B; \citet{meta2024introducing}), as well as Mistral AI's Mixtral-Instruct (8x7B; \citet{jiang2024mixtral}). 

We use default system prompts, zero-shot experimental setup, and greedy prediction mode, where temperature is set to 0. Our per-dataset prompts to LLMs are detailed in Appendix~\ref{sec:promts}.

\subsection{Experimental Results}

Table~\ref{tbl:main-results} shows per-dataset performance by model.  Original performance (average on examples in the raw dataset -- mean($score_i^o$)) is reported, as well as the average over mean perturbed sets -- mean($score_i^p$).  Figure~\ref{fig:metric_summary} in Appendix~\ref{sec:full_results} visualizes this table along with the 95\% confidence intervals for the metrics, represented by the cell shading in the table.  The significance thresholds in Table~\ref{tab:effect_size_thresholds} can tell us how large the observed mean $\NCoHfunc$ across examples in a dataset is, and one can decide that, say, a difference that is `small' or greater indicates the LLM is not robust to the perturbations.  The 95\% interval for $\NCoHfunc$ (or $\ANCoHfunc$) measure how variable this assessment of robustness is to random sampling; for instance, if the 95\% interval is itself also within the bounds of a `small' difference, this lends statistical confidence to the assessment that the LLM is robust.  The fact that $\NCoHfunc$ has such significance thresholds that can be used in a practical decision gives it a benefit compared to metrics like PDR, which lack them.

A slightly inferior performance on the perturbed instances compared to original is reflected by the negative Cohen's $h$ effect size ($\NCoHfunc$). Notably, the absolute value effect ($\ANCoHfunc$) differs considerably from the directional $\NCoHfunc$ values, indicating that the LLMs' predictions varied in both directions (better or worse) compared to the original instance performance -- a finding largely supportive of our assumption that naturally-occurring, non-malicious perturbations may have either positive or negative effect on a model's accuracy. While only a single observed $\NCoHfunc$ directional effect size is considered non-negligible (grey background), most absolute values constitute a "small" change, and virtually all effect sizes show significant, indicating small, yet systematic, and reliably detected change.

\input{chapters/benchmarking-results-table}

\begin{figure}[h!]
\centering
\includegraphics[width=0.95\columnwidth]{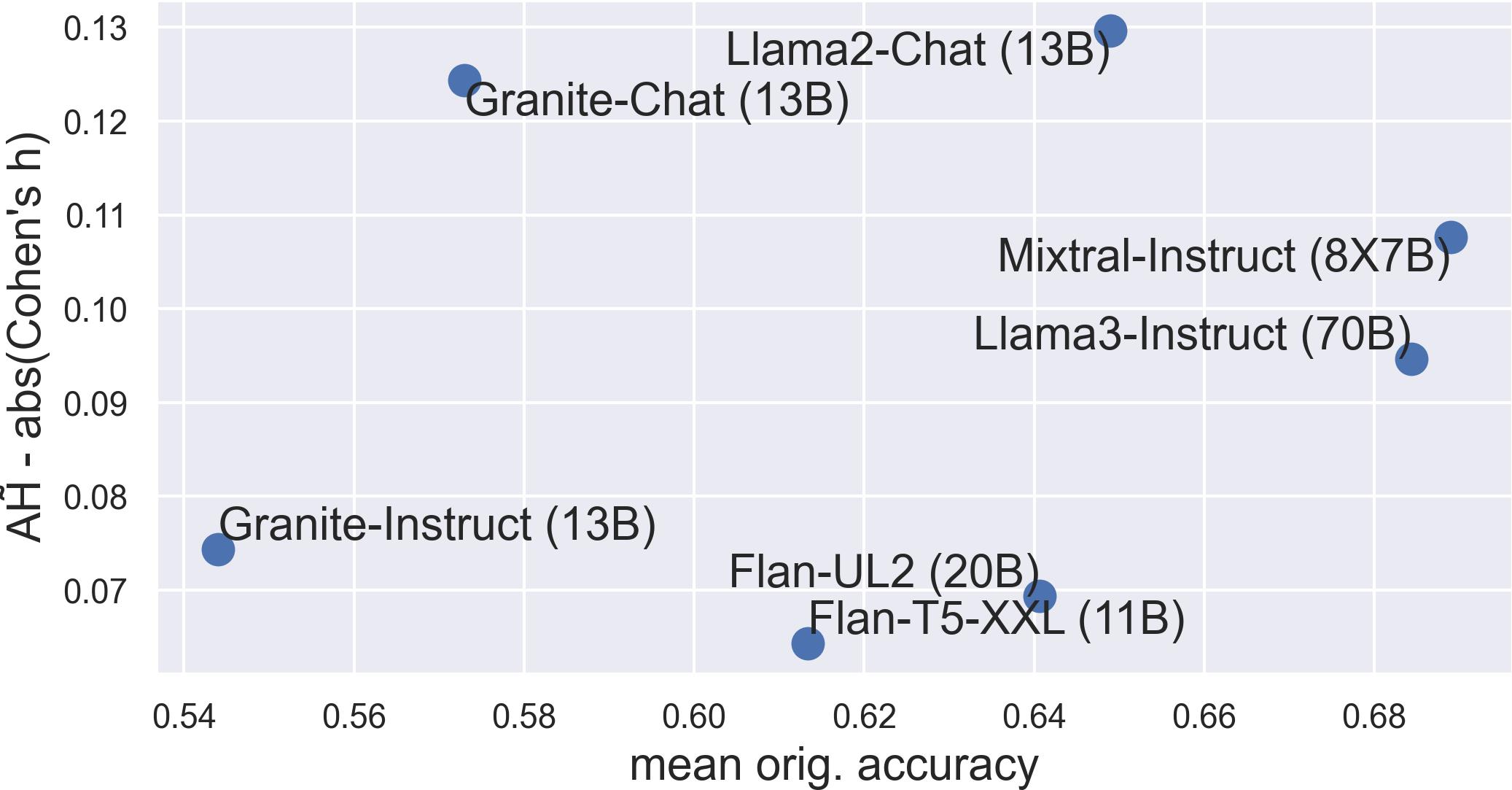}
\caption{Mean model accuracy on original datasets vs its undirectional robustness. x-axis: the higher, the better performing; y-axis: the lower, the more robust.}
\vspace{-0.1in}
\label{fig:perf-vs-robustness}
\end{figure}

\paragraph{Model Robustness vs Performance}
Figure~\ref{fig:perf-vs-robustness} illustrates average model accuracy on the original datasets vs their mean undirectional robustness ($\ANCoHfunc$). Evidently, best performing, recently released models (Llama3-Instruct (70B) and Mistral-Instruct (8{$\times$}7B)) exhibit more moderate robustness, compared to the most robust Flan-T5-XXL, that shows slightly inferior average performance. Granite-Instruct (13B) is one of the most robust models, while performing worse on average.

\paragraph{Robustness Evaluation by Perturbation Type}
We further break down the robustness measurements by individual perturbation types in Table~\ref{tbl:break-down-perturbation} for \texttt{Llama3-Instruct(70B)} -- one of the best-performing models on our datasets. Notably, paraphrasing the original question results in a more considerable performance drop than introducing superficial (simple) perturbations, across all three datasets. We attribute the particularly high absolute effect size in the PopQA dataset (0.15) to the fact that paraphrases in this dataset were created manually, aiming at high linguistic diversity while maintaining the original semantics.

%% file: chapters/benchmarking-results-table.tex
\begin{table*}[h!]
\resizebox{1.0\textwidth}{!}{
\begin{tabular}{l|cccl|cccc|cccc}
& \multicolumn{4}{c|}{PopQA} & \multicolumn{4}{c|}{BoolQ} & \multicolumn{4}{c}{SIGA} \\ 
model & M(orig) & M(pert.) & $\NCoHfunc$ & $\ANCoHfunc$ & M(orig) & M(pert.) & $\NCoHfunc$ & $\ANCoHfunc$ & M(orig) & M(pert.) & $\NCoHfunc$ & $\ANCoHfunc$ \\ \hline
Granite-Chat (13B)          & 0.20 & 0.18 & -0.02* & \cellcolor{gray!15}0.08* & 0.81 & 0.78 & -0.04* & \cellcolor{gray!15}0.11* & 0.71 & 0.67 & -0.04* & \cellcolor{gray!15}0.18* \\
Granite-Instruct (13B)      & 0.16 & 0.15 & -0.02\phantom{*} & \textbf{0.06}* & 0.87 & 0.86 & -0.02* & 0.06* & 0.60 & 0.59 & \textbf{-0.01}\phantom{*} & \cellcolor{gray!15}0.10* \\
Llama2-Chat (13B)           & 0.29 & 0.27 & -0.03* & \cellcolor{gray!15}0.12* & 0.84 & 0.81 & -0.04* & \cellcolor{gray!15}0.11* & \textbf{0.81} & 0.78 & \cellcolor{gray!15}-0.07* & \cellcolor{gray!15}0.16* \\
Llama3-Instruct (70B)       & 0.37 & 0.33 & -0.04* & \cellcolor{gray!15}0.15* & 0.89 & 0.87 & -0.03* & \cellcolor{gray!15}0.07* & 0.80 & \textbf{0.79} & \textbf{-0.01}\phantom{*} & \cellcolor{gray!15}\textbf{0.07}* \\
Mixtral-Instruct (8{$\times$}7B) & \textbf{0.39} & \textbf{0.36} & -0.04* & \cellcolor{gray!15}0.16* & 0.89 & 0.87 & -0.03* & \cellcolor{gray!15}0.07* & 0.79 & 0.78 & -0.02* & \cellcolor{gray!15}0.10* \\
Flan-T5-XXL (11B)           & 0.13 & 0.13 &  \textbf{0.00}\phantom{*} & \textbf{0.06} & 0.92 & 0.91 & \textbf{-0.02}* & 0.05* & 0.79 & 0.78 & -0.03* & \cellcolor{gray!15}0.09* \\
Flan-UL2 (20B)              & 0.15 & 0.14 & -0.01\phantom{*} & \cellcolor{gray!15}0.07* & \textbf{0.97} & \textbf{0.95} & -0.04* & \textbf{0.04}* & 0.80 & \textbf{0.79} & -0.02* & \cellcolor{gray!15}0.10* \\
\end{tabular}
}
\caption{Mean accuracy on the original datasets, mean accuracy on perturbed variants (slightly lower). Confidence intervals (CIs) for are calculated by original or perturbed group-level bootstrapping, as discussed in Appendix~\ref{sec:bootstrapping}. 
The high difference between the directional $\NCoHfunc$ and undirectional $\ANCoHfunc$ is suggestive of both increase and decrease in models' performance on original, compared to perturbed examples. Results for which the $\NCoHfunc$ and $\ANCoHfunc$ 95\% confidence intervals (see Appendix~\ref{sec:bootstrapping}.) do not contain 0 are marked with "*", indicating the significance of the finding. Notably, significant $\NCoHfunc$ and $\ANCoHfunc$ values may still indicate a very small effect size (see Appendix~\ref{sec:appendix_effect_size}); values reflecting a non-negligible change are marked with gray background. The best result in a column is boldfaced.}
\label{tbl:main-results}
\end{table*}

\begin{table*}[h!]
\centering
\resizebox{0.9\textwidth}{!}{
\begin{tabular}{l|c|ccc|ccc|ccc}
& & \multicolumn{3}{c|}{superficial (S)} & \multicolumn{3}{c|}{paraphrase (P)} & \multicolumn{3}{c}{distraction (D)} \\ 
dataset & M(orig) & M(pert.) & $\NCoHfunc$ & $\ANCoHfunc$ & M(pert.) & $\NCoHfunc$ & $\ANCoHfunc$ & M(pert.) & $\NCoHfunc$ & $\ANCoHfunc$ \\ \hline
PopQA   & 0.37 & 0.32 & -0.05* & \cellcolor{gray!15}0.12* & 0.34 & -0.03\phantom{*} & \cellcolor{gray!15}0.15* & -- & -- & -- \\
BoolQ   & 0.89 & 0.88 & -0.01* & 0.04* & 0.85 & -0.04* & \cellcolor{gray!15}0.07* & 0.88 & -0.01 & 0.05* \\
SIGA    & 0.80 & 0.79 & -0.01* & 0.06* & 0.77 & 0.00\phantom{*} & \cellcolor{gray!15}0.09* & -- & -- & -- \\
\end{tabular}
}
\caption{Mean accuracy on the original datasets, mean accuracy on perturbed variants with the most recent \texttt{Llama3-Instruct(70B)} model in this study, with break-down by variant type (superficial (S), paraphrase (P), distraction (D)). Results for which the $\NCoHfunc$ and $\ANCoHfunc$ 95\% confidence intervals (see Appendix~\ref{sec:bootstrapping}.) do not contain 0 are marked with "*", indicating the significance of the finding; values reflecting a non-negligible change are marked with gray background. Notably, paraphrasing the original question results in a more considerable performance drop than introducing superficial (simple) perturbations, across all three datasets.}
\label{tbl:break-down-perturbation}
\end{table*}

%% file: chapters/conclusion.tex
\section{Conclusions}
\label{sec:conclusions}

We evaluate the robustness of several LLMs on multiple diverse datasets, by expanding them with non-malicious, naturally-occurring perturbations, and measuring models' resilience to these variants in user input. We propose and evaluate a novel application of a statistical effect size metric for assessing model robustness in tasks with binary- or proportion- valued evaluation scores, and demonstrate its benefits in the non-adversarial scenario. 


%% file: chapters/limitations.tex
\section{Limitations}
\label{sec:limitations}

Our study, while contributing valuable insights for measuring model robustness to non-adversarial perturbations, is subject to several limitations. First, the application of Cohen's $h$ effect size, suggested is this work, is an intuitive fit for tasks with binary-valued evaluation outcome, correlating with PDR (denoting fractional decrease); other effect size metrics could constitute a more intuitive choice in scenarios with continuous evaluations scores. Second, a limited number of open models were evaluated on three datasets; the study can be extended to additional (commercial) models and more sophisticated tasks, e.g. MMLU \citep{hendryckstest2021}. Finally, while our automatic paraphrase generation is of high-quality overall, it is admittedly conservative -- only slight deviations from the original examples were applied to preserve semantics. We plan to make use of advanced models for more diverse paraphrase generation in the future.

%% file: chapters/appendix.tex
\appendix
\section{Appendix}

\subsection{Effect Sizes and Cohen's $h$}
\label{sec:appendix_effect_size}

Effect size metrics are measures of the size of a statistical phenomenon; common examples are Pearson correlation and odds ratios.  An effect size metric measures the aspect of interest (e.g., the difference in means between two sample sizes) in a way that is independent of the sample sizes. This is in contrast to the p-value of a hypothesis test statistic (e.g., a two-sample test) which, for fixed values of the sample means and variances, becomes more significant when the sample sizes increase \cite{sullivan2012using}; this quality makes p-values vulnerable to manipulation ("p-hacking").  Effect size metrics are often used to ensure that a hypothesis test has enough statistical power (complement of the Type-II or false negative error probability) given the sample size(s).
This insensitivity to the sample size in the effect size value means that they can be used to measure the significance of an effect in cases of small sample sizes (e.g., in our case when we have one original instance and a small number of perturbations), and thus may be better than, say, a two-sample p-value, for assessing robustness.  Cohen's $h$ has a particular advantage in that it is defined even if there is no sample variation (e.g., if $score\in\{0,1\}$), which causes p-values and some other effect sizes to be undefined.

Cohen's $h$---and thus $\NCoHfunc$---changes non-linearly with changes in $|score_i^o{-}score_i^p|$.  In contrast, PDR changes linearly when $score_i^o$ is fixed.  Considering the specific case when the original instance accuracy $score_i^o$ is perfect (1.0), both  $\NCoHfunc$ and "reverse PDR" (${=}-1{\times}\textrm{PDR}$) take values in the [-1, 0] range and are highly correlated, as shown in Figure~\ref{fig:pdr_cohensh}; this high correlation suggests that in this case ($score_i^o{=}1.0$), Cohen's $h$ constitutes an intuitive and easily-interpretable alternative to PDR.


A two-sample proportions difference hypothesis test (see \href{https://www.statsmodels.org/dev/generated/statsmodels.stats.proportion.proportions_ztest.html}{statsmodels' \texttt{proportions\_ztest}}, \citet{seabold2010statsmodels}) is an alternative way of measuring the significance of these differences.  In this test, the test statistic is maximized (i.e., is more significant) for a given fixed difference $|score_i^o{-}score_i^p|$ when one of the proportions is equal to 0 or 1, because the pooled variance in the denominator is minimized. The arcsine transformation in Cohen's $h$ magnifies the resulting effect size for a given $|score_i^o{-}score_i^p|$ when one of the proportions is close to 0 or 1 since the difference is more detectable, which causes the non-linear change in Figure~\ref{fig:pdr_cohensh} (around 0.1 and 0.9).

The corollary of the fact that Cohen's $h$ changes non-linearly with $|score_i^o{-}score_i^p|$ is that Cohen's $h$ (and $\NCoHfunc$) for pairs of $(score_i^o,\: score_i^p)$ should be equal when the difference is equally \textit{detectable} \cite[p. 180--181]{C1988}, despite $|score_i^o{-}score_i^p|$ differing.  Thus, for instance, $\NCoH{0.8}{ 1.0}{\approx}0.295$ but $\NCoH{0.6}{0.8}{\approx}0.141$, meaning that the 0.2 accuracy decrease from $score_i^o{=}1.0$ to $score_i^p{=}0.8$ should statistically be more than twice as detectable as the same decrease from $score_i^o{=}0.8$ to $score_i^p{=}0.6$.  Perturbation accuracy would have to fall from 0.8 to $score_i^p{\approx}0.36$ to be as significant, by $\NCoHfunc$, as the fall from 1.0 to 0.8, despite the raw decrease being more than twice as large.  

We note that if the instance scores $score_i^j$ are continuous-valued rather than binary, an alternative effect size metric such as Cohen's $d$ \cite{C1988} or Hedges' $g$ \cite{hedges1981distribution} for comparing sample means can be used in a similar way to Cohen's $h$. However, these effect sizes, unlike Cohen's $h$, are undefined or infinite when the within-sample variance is zero. We leave further investigation for future work.

\begin{table}[h!]
\centering
\resizebox{1.0\columnwidth}{!}{
\begin{tabular}{l|cc} 
effect size & Cohen's $h$ & $\NCoHfunc$ (normalized)\\ \hline
essentially zero & [0.0,\: 0.01) & [0.0,\: 0.0032) \\ 
very small & [0.01,\: 0.2) & [0.0032,\: 0.0637) \\
small & [0.2,\: 0.5) & [0.0637,\: 0.1592) \\
medium & [0.5,\: 0.8) & [0.1592,\: 0.2546) \\
large & [0.8,\: 1.2) & [0.2546,\: 0.3820)\\ 
very large & [1.2,\: 2.0) & [0.3820,\: 0.6366) \\ 
huge & [2.0,\: $\pi$] & [0.6366,\: 1.0] \\  
\end{tabular}
}
\caption{Ranges of values of Cohen's $h$ and their size interpretation, as defined by \citet{C1988} and \citet{sawilowsky2009new}; many other related metrics, such as Cohen's $d$, have the same thresholds but are not bounded from above.  The bounds for our normalized metric ($\NCoHfunc$) are the first bounds, divided by $\pi$.}
\label{tab:effect_size_thresholds}
\end{table}

\subsection{Bootstrapped Confidence Intervals}
\label{sec:bootstrapping}

All metrics and instance scoring functions are implemented in the open-source repository \texttt{unitxt} \cite{bandel2024unitxt}.  A `group' here consists of an original instance and its $m(i)$ perturbations (see Section~\ref{ssec:input_perturbations}).  A given metric $f$ (e.g., mean score, PDR, $\NCoHfunc$) produces a group-level score $s_i{=}f\left(score_i^o, score_i^1,\dots,score_i^{m(i)}\right)$. Thus, if original dataset $\mathcal{D}$ has $n$ instances, we have $n$ instance-group scores $(s_1,\dots,s_n)$ on $\mathcal{D}'$. Statistical analysis of the metric is done by constructing 95\% bootstrapped confidence intervals on the $s_i$ scores, discarding any undefined values, rather than resampling the instance scores, reforming the groups, and calculating the group scores; the latter option could result in duplicated perturbed instances or incomplete groups if $score_i^o$ is missing, which can make $s_i$ undefined.  We conduct group-score resampling because we analyze the typical average robustness across original instances, and thus the unit of analysis is the original instance \emph{together with} its perturbations, as reflected in $s_i$.

\subsection{Detailed Experimental Results}
\label{sec:full_results}

\begin{figure*}[h!]
\centering
\includegraphics[width=0.80\textwidth]{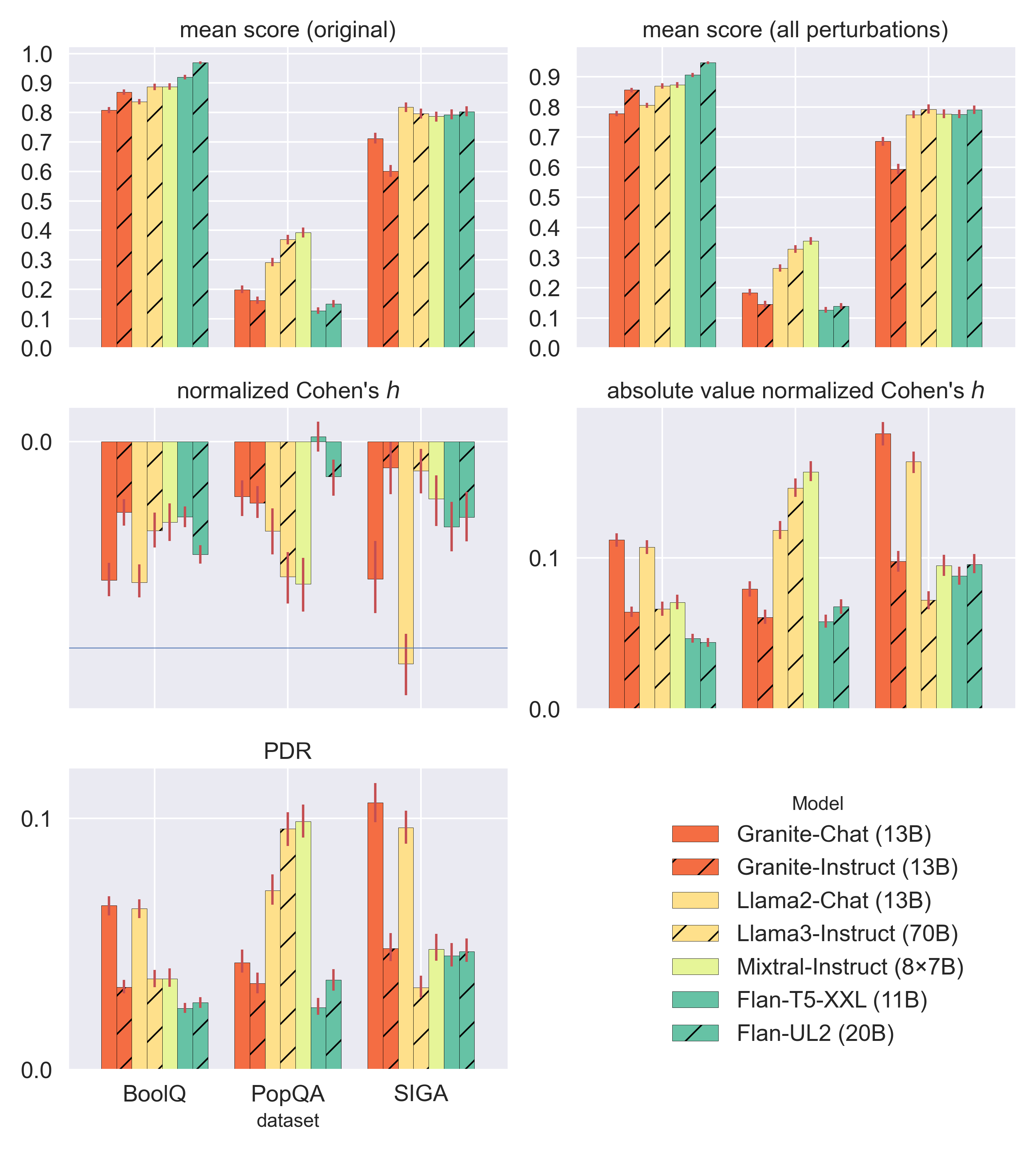}
\caption{Mean metric scores by model and dataset. Red error bars show a 95\% bootstrapped confidence interval.}
\label{fig:metric_summary}
\end{figure*}

Figure~\ref{fig:metric_summary} shows the robustness metric values by model and dataset.  The bar colors represent the model source (Google, IBM, Meta, Mistral), while the individual models are distinguished by the diagonal hashing pattern.  At the top of each bar, a red line shows the 95\% bootstrapped confidence interval as described in Appendix~\ref{sec:bootstrapping}.

Note that the values of normalized Cohen's $h$ (i.e., $\NCoHfunc$) are almost always negative, indicating a decrease in accuracy after perturbation; however, these changes are nearly all very minor, falling between 0 and the "very small" decrease threshold of ${\approx}-0.0637$, shown by the horizontal blue line.  These insignificant changes in performance are reflected in the fact that the heights of the mean original score $score_i^o$ bars (top left) are very similar to the corresponding mean perturbed score $score_i^p$ bars (top right), and that their confidence interval bands overlap significantly.  Furthermore, the relative magnitudes of $\NCoHfunc$ and $\ANCoHfunc$ also correlate well with PDR, which, despite its drawbacks (see Section~\ref{sssec:pdr}), can serve as a good sanity check.

\subsection{Experimental Setup and Prompts}
\label{sec:promts}

We use default system prompts, zero-shot experimental setup, and greedy prediction mode, where the temperature is set to 0. Our per-dataset prompts for the models are detailed below. 

\setlist[itemize]{leftmargin=*}

\begin{itemize}
\item \textbf{BoolQ}:\\
\texttt{"Context: \{passage\}\\Based on the context, answer the following question with a single word: yes or no. Question: \{question\}\\Answer:"}

\item \textbf{PopQA}:\\
\texttt{"Question: \{question\}\\Answer:"}

\item \textbf{SIGA}: \\
\texttt{"Does the following statement contain abusive or unfair negative opinion against some social identity group? Answer with yes or no.\\Statement: \{question\}\\Answer:"}

\end{itemize}